\documentclass[sigconf]{acmart}
\usepackage{amsmath}
\usepackage{pifont}
\AtBeginDocument{%
  }

\copyrightyear{2025}
\acmYear{2025}
\setcopyright{acmlicensed}\acmConference[WWW Companion '25]{Companion Proceedings of the ACM Web Conference 2025}{April 28-May 2, 2025}{Sydney, NSW, Australia}
\acmBooktitle{Companion Proceedings of the ACM Web Conference 2025 (WWW Companion '25), April 28-May 2, 2025, Sydney, NSW, Australia}
\acmDOI{10.1145/3701716.3715204}
\acmISBN{979-8-4007-1331-6/25/04}

\begin{document}

\title{Intelligent Legal Assistant: An Interactive Clarification System for Legal Question Answering}


\author{Rujing Yao}
\affiliation{%
  \institution{Nankai University}
  \city{Tianjin}
  \country{China}}
\email{rjyao@mail.nankai.edu.cn}

\author{Yiquan Wu}
\affiliation{%
  \institution{Zhejiang University}
  \city{Hangzhou}
  \country{China}}
\email{wuyiquan@zju.edu.cn}

\author{Tong Zhang}
\affiliation{%
  \institution{Nankai University}
  \city{Tianjin}
  \country{China}}
\email{tongzhangnk@mail.nankai.edu.cn}

\author{Xuhui Zhang}
\affiliation{%
  \institution{Fayuan Inc.}
  \city{Hangzhou}
  \country{China}}
\email{zhangxh999@gmail.com}

\author{Yuting Huang}
\affiliation{%
  \institution{Zhejiang University}
  \city{Hangzhou}
  \country{China}}
\email{yutinghuang@zju.edu.cn}

\author{Yang Wu}
\affiliation{%
  \institution{Worcester Polytechnic Institute}
  \city{Worcester}
  \country{USA}}
\email{ywu19@wpi.edu}

\author{Jiayin Yang}
\affiliation{%
  \institution{Zhejiang University}
  \city{Hangzhou}
  \country{China}}
\email{jiayinyang18@gmail.com}

\author{Changlong Sun}
\affiliation{%
  \institution{Zhejiang University}
  \city{Hangzhou}
  \country{China}}
\email{changlong.scl@gmail.com}

\author{Fang Wang}
\affiliation{%
  \institution{Nankai University}
  \city{Tianjin}
  \country{China}}
\email{wangfangnk@nankai.edu.cn}

\author{Xiaozhong Liu}
\authornote{Corresponding author.}
\affiliation{%
  \institution{Worcester Polytechnic Institute}
  \city{Worcester}
  \country{USA}}
\email{xliu14@wpi.edu}

\renewcommand{\shortauthors}{Rujing Yao et al.}

\begin{abstract}
The rise of large language models has opened new avenues for users seeking legal advice. However, users often lack professional legal knowledge, which can lead to questions that omit critical information. This deficiency makes it challenging for traditional legal question-answering systems to accurately identify users' actual needs, often resulting in imprecise or generalized advice. In this work, we develop a legal question-answering system called Intelligent Legal Assistant, which interacts with users to precisely capture their needs. When a user poses a question, the system requests that the user select their geographical location to pinpoint the applicable laws. It then generates clarifying questions and options based on the key information missing from the user's initial question. This allows the user to select and provide the necessary details. Once all necessary information is provided, the system produces an in-depth legal analysis encompassing three aspects: overall conclusion, jurisprudential analysis, and resolution suggestions. More materials of the system can be accessed at \url{https://github.com/RujingYao/Intelligent-Legal-Assistant}.
\end{abstract}




\maketitle
\section{INTRODUCTION}
The demand for legal services has seen exponential growth, escalating the importance of legal question-and-answer (Q\&A) systems as pivotal resources for legal professionals and the general public alike. These systems offer swift, precise legal information, playing a vital role in democratizing access to legal knowledge. By making legal information more accessible, they empower non-specialists who might not have the financial resources or opportunity to consult a lawyer directly. Despite these advancements, the specialized nature of legal terminology and procedures remains a significant barrier. This complexity often prevents the general public from possessing the necessary professional knowledge to accurately communicate essential information when seeking legal advice. Such a gap in understanding can hinder the effectiveness of legal Q\&A systems, as these platforms struggle to comprehend the true needs of their users, thereby complicating the delivery of accurate and customized legal advice.

\begin{figure}[t]
\setlength{\belowcaptionskip}{-0.3cm}
    \centering
    \includegraphics[width=0.9\linewidth]{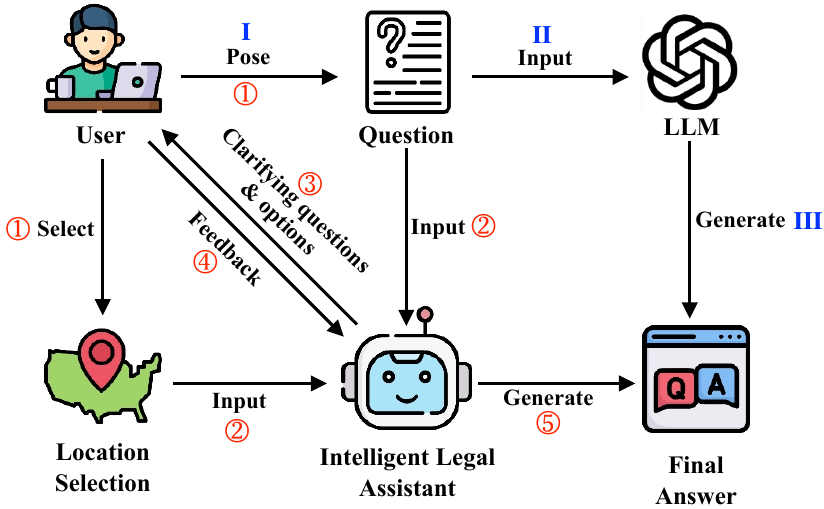}
    \caption{Comparison of our system with traditional LLM-based Q\&A systems. \ding{172}-\ding{176} are the steps of our system, and I-III are the steps of traditional LLM-based Q\&A systems.}
    \label{figure1}
\end{figure}

\begin{figure*}[!h]
\setlength{\belowcaptionskip}{-0.3cm}
    \centering
    \includegraphics[width=0.8\linewidth]{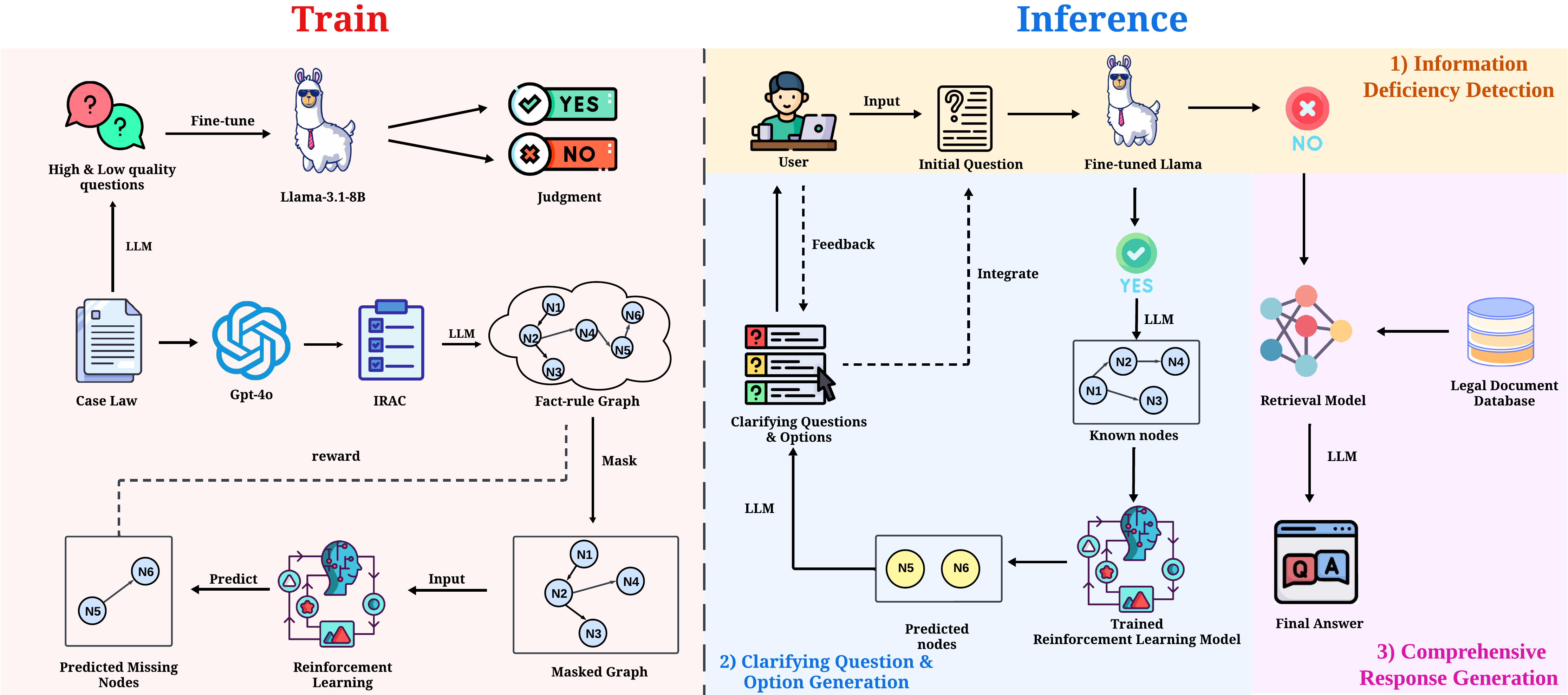}
    \caption{System architecture. The system consists of three primary functional modules: 1) information deficiency detection, 2) clarifying question and option generation, and 3) comprehensive response generation.}
    \label{framework}
\end{figure*}

Recent developments in Large Language Models (LLMs) have introduced numerous legal Q\&A platforms, such as AI Lawyer~\footnote{https://ailawyer.pro/} and Callidus AI~\footnote{https://callidusai.com/}. These systems, however, primarily rely on user-submitted questions to generate responses. If the input from the user lacks critical information, the ability of these systems to furnish comprehensive and relevant advice is notably diminished. To address this issue, this study develops an innovative interactive legal question and answer system built upon LLMs, as shown in Figure~\ref{figure1}. Traditional LLM systems typically involve a user posing a question, with the LLMs then providing a response. We propose Intelligent Legal Assistant, which not only asks the user to pose a question but also requires them to submit information about their geographical location to effectively provide legal assistance based on the law. 

The proposed system initially assesses whether the user's question contains all the necessary details. If it does, the system proceeds to generate an immediate response. Conversely, if the query is incomplete, the system employs a novel approach by generating targeted clarifying questions and presenting options to the user. This process ensures that all requisite information is gathered before formulating a legal response. After the user addresses these clarifying questions, our system evaluates the completeness of the information. Once deemed complete, a professional legal response is crafted based on the thoroughly formed user question. The main contributions of our work include 1) the development of a method to identify missing information, 2) the generation of precise clarifying questions, and 3) the integration of domain-specific knowledge. This approach allows our system to offer more accurate and professionally relevant legal advice, thus enhancing the utility and effectiveness of legal Q\&A systems in practice.

\section{SYSTEM OVERVIEW} \label{overview}
Figure~\ref{framework} illustrates the system architecture, which consists of three primary functional modules: 1) information deficiency detection, 2) clarifying question and option generation, and 3) comprehensive response generation.

\subsection{Information Deficiency Detection}
Since case law data comprises complete legal cases, GPT-4o is employed to generate questions of varying quality from this data, including low-quality questions that lack key information and high-quality questions that are fully informative. Low-quality questions are information-deficient and thus require the generation of clarifying questions to fill in these gaps, whereas high-quality questions, which are complete, do not necessitate further inquiry. Inspired by~\citet{schick2024toolformer} and~\citet{wu2024knowledge}, we implement prompt-based fine-tuning on the Llama-3.1-8B model. The reason is that Llama-3.1-8B provides a good balance between computational efficiency and the ability to handle the depth of understanding required for legal texts. It is large enough to perform complex tasks but not as resource-intensive as larger models, making it suitable for applications where frequent inference is needed, such as in real-time legal question analysis. Upon receiving a legal question input, the Llama-3.1-8B model evaluates whether information is lacking. It outputs `yes' if there is a deficiency, indicating that further clarification is required, and `no' otherwise.

\subsection{Clarifying Question and Option Generation}

\subsubsection{Construction of key fact-rule node graph}\label{sec1}

A complete legal case can be analyzed as a graph consisting of multiple key fact-rule nodes. To accurately capture these nodes, this study parses each case law document into an IRAC (Issue, Rule, Application, Conclusion) structure to extract critical information. From each IRAC framework, a fact-rule node graph $g_i$ is extracted. Ultimately, these graphs from all case law documents are merged to form the final fact-rule node graph, denoted as $\mathcal{G}$. For each IRAC framework, we utilize a LLM to summarize it into a comprehensive question, denoted as $q_i$, and its corresponding set of key nodes is denoted as $N_i$. We randomly mask some nodes within $N_i$, using the LLM to generate the masked version of the problem $\hat{q}_i$, with the masked nodes denoted as $\hat{N}_i$.

\subsubsection{Missing Node Prediction}
Given a user's question, our goal is to train a model capable of predicting missing key elements. Due to the complexity and variability of legal issues, traditional supervised learning methods may not effectively capture unforeseen structures in real-world scenarios. To address this challenge, we utilize reinforcement learning, which allows the model to independently explore and learn within either a simulated or an actual interactive environment. This capability is particularly crucial for adapting to frequently emerging and rare situations in the legal field. The policy gradient method enables the model to learn how to predict key elements of problems through trial and error, which is highly suitable for the high uncertainty and complex decision-making processes typical of legal issues. This method also optimizes long-term rewards directly, facilitating the development of more comprehensive and robust prediction strategies. Deep Deterministic Policy Gradient (DDPG)~\citep{lillicrap2015continuous} is employed in this study to predict missing nodes, utilizing the Actor-Critic framework.

Initially, we employ a LLM to identify nodes in graph \( G \) relevant to the user's question $q_i$. These nodes, denoted as $X_{\text{known}} = \{x_j\}_{j=1}^n$, where $c_j$ represents the $j$-th node, are subsequently processed using a Graph Neural Network (GNN)~\citep{scarselli2008graph} to obtain their embeddings. GNN is well-suited for learning on graph-structured data. In the legal domain, where relationships and dependencies between different elements are crucial, GNN effectively captures these structural details. This allows the model to understand how different legal concepts are interconnected within graph $\mathcal{G}$.
\begin{equation}
E_{x_j} = \text{Encode}(x_j),
\end{equation}
where $E_{x_j}$ represents the embedding vector of  $x_j$.

Then, a policy network \( \pi(a_t|s_t) \) and a value network \( V_\phi(s_t) \) are employed to predict the next node \( a_t \) and to evaluate the cumulative reward at \( s_t \). The state \( s_t \) is defined by the set of currently known nodes \( X_{\text{known}} \). The state \( s_t \) is iteratively updated by appending each predicted node \( a_t \) to \( X_{\text{known}} \), leading to the dynamic updating of the state representation.

The training objective is to maximize the expected cumulative reward, and the integral of the reward function over the policy is calculated as follows:
\begin{equation}
J(\theta) = \mathbb{E}\left[\sum_{t=0}^{\infty} (\gamma^t \cdot (r_t + \gamma V_\phi(s_{t+1}) - V_\phi(s_t)))\right],
\end{equation}

where \( r_t \) is the immediate reward received at time \( t \); \( \gamma \) is the discount factor used to compute the present value of future rewards, ranging between 0 and 1; \( V_{\phi}(s_{t+1}) \) and \( V_{\phi}(s_t) \) are the value functions, parameterized by \(\phi\), which evaluate the expected total returns at states \( s_{t+1} \) and \( s_t \).

\subsubsection{Generation of clarifying questions and options}

The trained model described in Section~\ref{sec1} assesses the user's question to identify missing key nodes. We then employ an in-context learning methodology, wherein the user's question along with the identified missing nodes is input into a LLM. This enables the LLM to generate targeted clarifying questions  \( C_{i,j} \)  and corresponding options, where \( C_{i,j} \)  represents the $j$-th clarifying question for the $i$-th user's question, thus enhancing the interaction by effectively addressing information gaps.

\begin{figure*}[ht]
\setlength{\belowcaptionskip}{-0.3cm}
    \centering
    \includegraphics[width=0.81\linewidth]{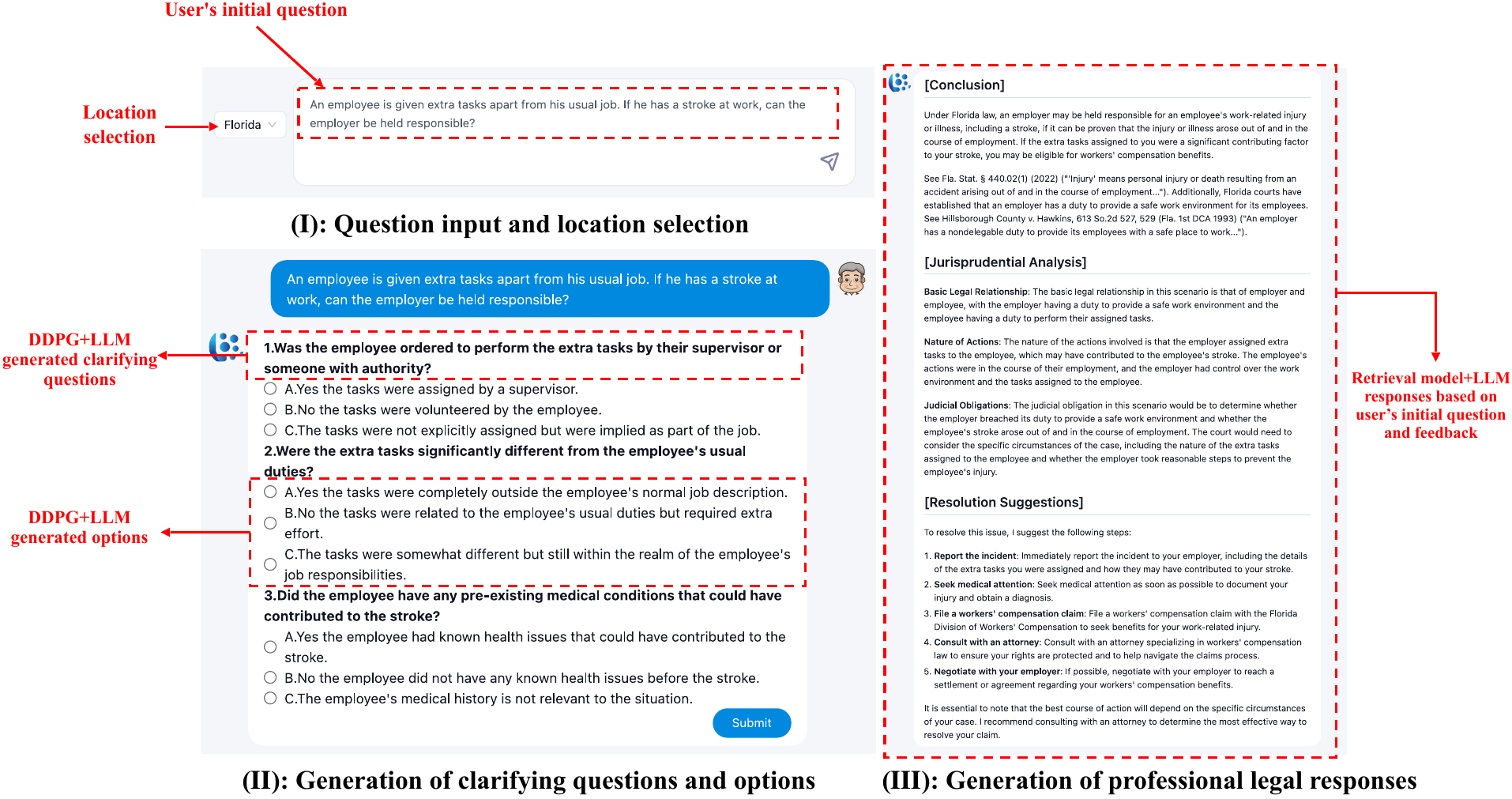}
    \caption{System Demo with User-LLM Interactions.}
    \label{demo}
\end{figure*}

\subsection{Comprehensive Response Generation}

After the clarifying question \( C_{i,j} \) is generated, the user must select an option for each clarifying question based on their specific circumstances. The selection is recorded as \( u_{ijk} \), indicating the \( k \)-th option for the \( j \)-th clarifying question of the \( i \)-th user's question. The clarifying question \( C_{i,j} \), the user's selection \( u_{ijk} \), and the user's initial question \( q_i \) are then provided to a traditional retrieval model to identify the most relevant laws from the legal database, represented as \( L_i = \{l_{i,1}, l_{i,2}, \ldots, l_{i,m}\} \). Specifically, the clarifying question \( C_{i,j} \), the user's selection \( u_{ijk} \), and the user's initial question \( q_i \) are concatenated to form a combined text, denoted as \( z_i \). This text is transformed into vector embeddings using the text-embedding-3-large model. Similarly, each legal document in the legal database is also embedded using the same model. Subsequently, the cosine similarity between the embedding of \( z_i \) and the legal document embeddings is calculated to retrieve the most relevant legal laws. This information is then provided to the LLM to generate the final answer.
\begin{equation}
A_i^\prime  = LLM(q_i,C_{i,j}, u_{ijk}, L_i).
\end{equation}

\section{DEMONSTRATION}

This demonstration describes an interactive system designed to provide the accuracy and specificity of responses to user questions, utilizing the techniques detailed in Section~\ref{overview}. The three main steps are illustrated in Figure~\ref{demo}.

\textbf{Initial input:} Our system prompts users to specify their location upon posing a legal question. This localization enables the provision of solutions that are pertinent to the legal standards and situational realities of different regions, thereby effectively addressing the users' legal requirements.

\textbf{Selection of options for clarifying questions:} Upon receiving a legal inquiry, the system assesses whether the question has informational deficiencies. If the inquiry is comprehensive, an immediate response is provided. If essential details are missing, the system generates targeted clarifying questions along with multiple-choice options that reflect these gaps. This approach minimizes misunderstandings due to insufficient information and precisely identifies the user’s specific needs. Subsequently, users select the options that best represent their actual situations.

\textbf{Professional legal responses:} By integrating the initial inquiry with the user’s answers to the clarifying questions, the system evaluates the completeness of the information provided. If the information is incomplete, the system offers additional clarifying questions and options. Once the information is deemed complete, the system produces an in-depth legal analysis encompassing three aspects: overall conclusion, jurisprudential analysis, and resolution suggestions.

\section{EVALUATION}
We invited 100 users to conduct a blind evaluation, comparing outputs from our system, GPT-4o, AI Lawyer, and Callidus AI. They rated each on three dimensions: accuracy, satisfaction, and usage preference. Accuracy refers to the degree to which the system provides correct or valid responses to the legal issues presented. Satisfaction measures the users' overall contentment with their interaction with the system. This encompasses various aspects of user experience, including ease of use, relevance and clarity of the information provided, and the extent to which the system meets the users' expectations and needs. Participants scored accuracy and satisfaction on a scale from 1 to 5, where 1 is the lowest and 5 is the highest. Regarding usage preference, users were asked which model they would prefer to use for resolving their legal issues.

The survey results are presented in Table~\ref{tab:human_evaluation}. They show that our system significantly outperformed other models in all measured aspects.

\begin{table}[htbp]
\centering
\small
\begin{tabular}{lccc}
\toprule
Method       & Accuracy & Satisfaction & Preference \\
\midrule
GPT-4o & 3.2  & 2.9 & 3\% \\
AI Lawyer & 3.7  & 3.2 & 2\% \\
Callidus AI & 3.8 & 3.6 & 5\% \\

Our system & 4.8  & 4.8 & 90\%  \\
\bottomrule
\end{tabular}
\caption{Results of human evaluation.}\vspace{-0.2in}
\label{tab:human_evaluation}
\end{table}

\section{CONCLUSION}

This paper introduces an innovative legal Q\&A system that leverages the capabilities of LLMs to interact with users through a series of clarifying questions, thereby compensating for the typical lack of legal knowledge among the general public. By engaging users in this dynamic dialogue, our system not only identifies and fills information gaps, but also tailors its advice to the specific needs of each user, enhancing the precision and relevance of the legal guidance provided. Additionally, the system includes a button that allows users to select their geographic location, offering more personalized and region-specific legal advice. Surveys have demonstrated that users are highly satisfied with the system and are eager to keep using it.

\section{Ethical Use of Data and Informed Consent}
The standards for data usage are strictly adhered to. All participants were thoroughly informed about the study's purposes, the nature of their involvement, and the use and confidentiality of the data collected.

\bibliographystyle{ACM-Reference-Format}
\balance
\bibliography{sample-base}










\end{document}